\documentclass[twocolumn,conference]{IEEEtran}
\usepackage[T1]{fontenc}
\usepackage[latin9]{inputenc}
\usepackage{graphicx}
\usepackage[unicode=true,
 bookmarks=true,bookmarksnumbered=true,bookmarksopen=true,bookmarksopenlevel=1,
 breaklinks=false,pdfborder={0 0 0},pdfborderstyle={},backref=false,colorlinks=false]
 {hyperref}
\hypersetup{pdftitle={Your Title},
 pdfauthor={Your Name},
 pdfpagelayout=OneColumn, pdfnewwindow=true, pdfstartview=XYZ, plainpages=false}

\makeatletter

\providecommand{\tabularnewline}{\\}

\usepackage[caption=false,font=footnotesize]{subfig}

\makeatother

\begin{document}
\title{Bag of Tricks for Retail Product Image Classification}
\author{\IEEEauthorblockN{Muktabh Mayank Srivastava}\IEEEauthorblockA{ParallelDots, Inc.\\
Email: muktabh@paralleldots.com}}
\maketitle
\begin{abstract}
Retail Product Image Classification is an important Computer Vision
and Machine Learning problem for building real world systems like
self-checkout stores and automated retail execution evaluation. In
this work, we present various tricks to increase accuracy of Deep
Learning models on different types of retail product image classification
datasets. These tricks enable us to increase the accuracy of fine
tuned convnets for retail product image classification by a large
margin. As the most prominent trick, we introduce a new neural network
layer called Local-Concepts-Accumulation (LCA) layer which gives consistent
gains across multiple datasets. Two other tricks we find to increase
accuracy on retail product identification are using an instagram-pretrained
Convnet and using Maximum Entropy as an auxiliary loss for classification.
\end{abstract}

\IEEEpeerreviewmaketitle{}

\section{Introduction}

Retail product image classification is the problem of deciphering
a retail product from its image. This recognition of products from
images is needed in a lot of Computer Vision applications in the real
world like self-checkout shops, retail execution measurement and shopper
behavior observation. Convnets (Convolutional Neural Networks) have
been shown to give the best performance for many image classification
datasets. Transfer Learning is a method to train a Deep Learning model
on a small dataset by finetuning a model pretrained on a larger dataset.
This practice is especially more prevalent with convnets which respond
very well to this method. This work aims to figure out the best method
to finetune deep convnets for different types of retail product classification
datasets.

Classifying and identifying retail product images is a very important
component of systems where one needs to automate or analyze retail
practices. It can help in making a self-checkout store by providing
an interface to recognize products for automatic billing, help automate
retail supply chain by automating product logging, can help automatically
evaluate retail-execution evaluation when combined with a retail product
object detector or help analyze consumer behavior in retail stores
in combination with a video analysis system. 

Deep Learning algorithms have gathered interest recently due to their
performance and applicability in the real world\cite{1}. Convolutional
Neural Networks (convnets), a type of Deep Learning algorithm has beaten the
state of the art results for various Computer Vision tasks like image
classification\cite{2}, object detection\cite{3} and image matching\cite{4}.
In scenarios like identifying retail products, where often only a
relatively small number of training images per class are available
(sometimes just a few product packshots\cite{7}), finetuning\cite{5}
of pre-trained weights is generally the preferred mode to train the
convnet in use. Few shot classification techniques\cite{6} are often
used in combination with finetuning when only a few images per class
are present. Our aim in this work is to come up with a set of tricks
which give high accuracy across different retail product classification
datasets when finetuning convnets. 

Our contributions in this work are : 1. We introduce Local-Concepts-Accumulation
layer, which gets consistent accuracy gains across datasets, 2. We
show that Maximum-Entropy loss\cite{6} can be used as an auxiliary loss in
combination with Local-Concepts-Accumulation to increase the classification
accuracy even more, 3. We show that a model of the exact same size
pretrained on Instagram and then imagenet\cite{8} gives better accuracy
than a model pretrained just on Imagenet.

\section{Related Work}
Deep Learning\cite{1} systems have been making inroads into many cognitive automation tasks. In case of retail, there is a lot of scope to make existing workflows efficient using Deep Learning. Localizing and classifying retail objects on retail shelves has been studied in the past. Traditional Image Processing features SIFT\cite{17} and Harris corners\cite{18} have been used for detecting and identifying retail products. \cite{16} proposes using SIFT features\cite{17} with a hybrid approach combining SVM with HMM/CRF for context aware product detection and identification. \cite{13} introduces Grozi-120 dataset and uses SIFT\cite{17} for product identification as baseline. \cite{14} introduces CAPG-GP dataset and uses a combination of Deep Learning\cite{5} and SIFT\cite{17}/BRISK\cite{23} for product recognition. \cite{19} compares visual bag of words and deep learning on grozi dataset for both detection on shelves and classification. \cite{20} tries out dense pixel based matching, bag of words and genetic algorithms for exemplar based product matching. \cite{21} uses Object Detection algorithms for one-shot product detection and identification. \cite{22} uses BRISK features\cite{23} and graphs to verify planograms. \cite{24} uses GAN based training of convnet embeddings for fine-grained product image classification.

With the recent introduction of generic retail object detectors from retail shelves like \cite{3} and \cite{25}, the problem of retail object detection from shelves and retail object identification can be separately solved. In this work, we show methods to improve finetuning\cite{5} of convnets for image classification of retail objects.

Convnets have been shown to work very well for image classification problems\cite{10}. It has also been shown that models pretrained on imagenet dataset can be finetuned on other smaller datasets\cite{5} for classification to achieve better accuracy. ResNext architecture\cite{2} is one of the best-performing deep learning architectures for image classification. ResNext architecture has residual connections with bottleneck dimensions between layers and also has multiple paths within each layer. Resnext architecture when trained on a larger instagram dataset and finetuned on imagenet, gives state of the art accuracy on imagenet\cite{8}. It is shown, such a network trained on instagram, also called ResNext-WSL\cite{9} is very robust to image noise and perturbations\cite{11}.

Traditionally, image matching techniques using descriptors like BRISK\cite{23} and SIFT\cite{17} have also been used to identify retail products. Superpoint\cite{4} is a convnet based keypoint detection and keypoint matching algorithms that has recently shown better results than SIFT. Superpoint is trained on a synthetic dataset and then finetuned on image augmentations to become invariant to various distortions. We use Superpoint as a baseline in all benchmark tasks. 

Fine-grained classification is a classification task where classes of images are visually similar to each other. Maximum Entropy Loss\cite{6} has been used to make classification in such scenario more effective. We use Maximum Entropy loss as we find the 'lack of diversity of features' hypothesis true in all retail product image classification, just like it is true in fine-grained classification. 

\section{Methodology}

We present three tricks which make convnets more effective at recognizing
retail product images. We try this on three different datasets which
represent different scenarios which arise while working with retail
images in real world. First trick we present is that a convnet\cite{8}
pretrained on instagram\cite{9} images and then finetuned later on
imagenet works better for retail product classification than a convnet
pretrained on imagenet\cite{10} alone. The second trick is the new
type of neural network layer called Local-Concepts-Accumulation we
introduce, which can be used while finetuning convnets for retail
product image classification. Local-Concepts-Accumulation layer is
a simple layer which can be added while finetuning any off-the-shelf
convnet. The third trick is to train a multitask learning based classifier
using Maximum Entropy loss as an auxiliary loss. We show that these
tricks incrementally give good gains for retail product image classification.

\subsection{Finetuning an Instagram pretrained model}

ResNext\cite{2} architecture has shown state of the art classification
accuracy on classification tasks in the past. We thus chose it as
our baseline architecture to finetune for retail product image classification.
It has been shown that a ResNext model pretrained on instagram\cite{9}
images using hashtags as training labels before finetuning it on imagenet\cite{10}
gives state of the art results on imagenet classification\cite{8}.
This convnet, also known as ResNext-WSL model, has been shown to have
more robustness on common image corruptions and perturbations\cite{11}.
We show that using a pretrained ResNext-WSL with the same number of
parameters gives better accuracy than a pretrained ResNext on imagenet
(we refer this ResNext model pretrained just on imagenet as ResNext-INet
henceforth to contrast it with ResNext-WSL). We use the \textquotedblleft resnext-101\_32X8\textquotedblright{}
pretrained models (both ResNext-INet and ResNext-WSL) available from
pytorch\cite{12} repositories for finetuning. Both the networks require
exactly the same resources for finetuning as they have the same number
of parameters and we finetune them with exactly the same hyperparameters.
We find that finetuning ResNext-WSL gets gains in accuracy for all
datasets we work on. It seems that ResNext-WSL gets better gain in
accuracy on datasets where test images have more noise and distortions. 

\subsection{Local-Concepts-Accumulation layer}

We introduce a new layer called Local-Concepts-Accumulation (LCA layer)
which can be added to any convnet architecture as its penultimate layer while
training or finetuning. We show that adding LCA layer while finetuning
convnets for retail product image classification gives sizable gains
in accuracy across all datasets we present our results on. The hypothesis
to use this layer is that there are multiple large or small \textquoteleft local
concepts\textquoteright{} in retail product images which when individually
recognized and then aggregated can be used to classify the product.

\begin{figure}[htbp]
\includegraphics[width=1\linewidth]{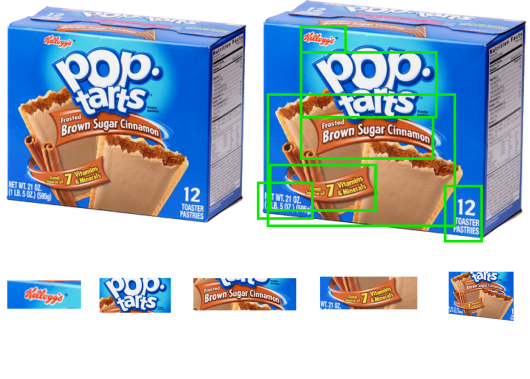}

\caption{Figure showing hypothesized local concepts on a retail product. The
top left image is of the retail object itself. The top right image
is of various possible local concepts marked on image. The bottom
row has images showing all the possible local concepts shown individually.}
\end{figure}

The hypothesis further is that when a convnet is trained to classify
pooled features from the last layer, it focuses more on the global
look and feel of the retail product rather than on the local features.
When we use different local concepts as features and aggregate their
contribution with equal importance, the classifier focuses on both
individual local concepts and global look and feel. It is proposed
that focusing on local concepts would give a boost in classification
accuracy.

\begin{figure}[htbp]
\begin{centering}
\textsf{\includegraphics[width=1\linewidth]{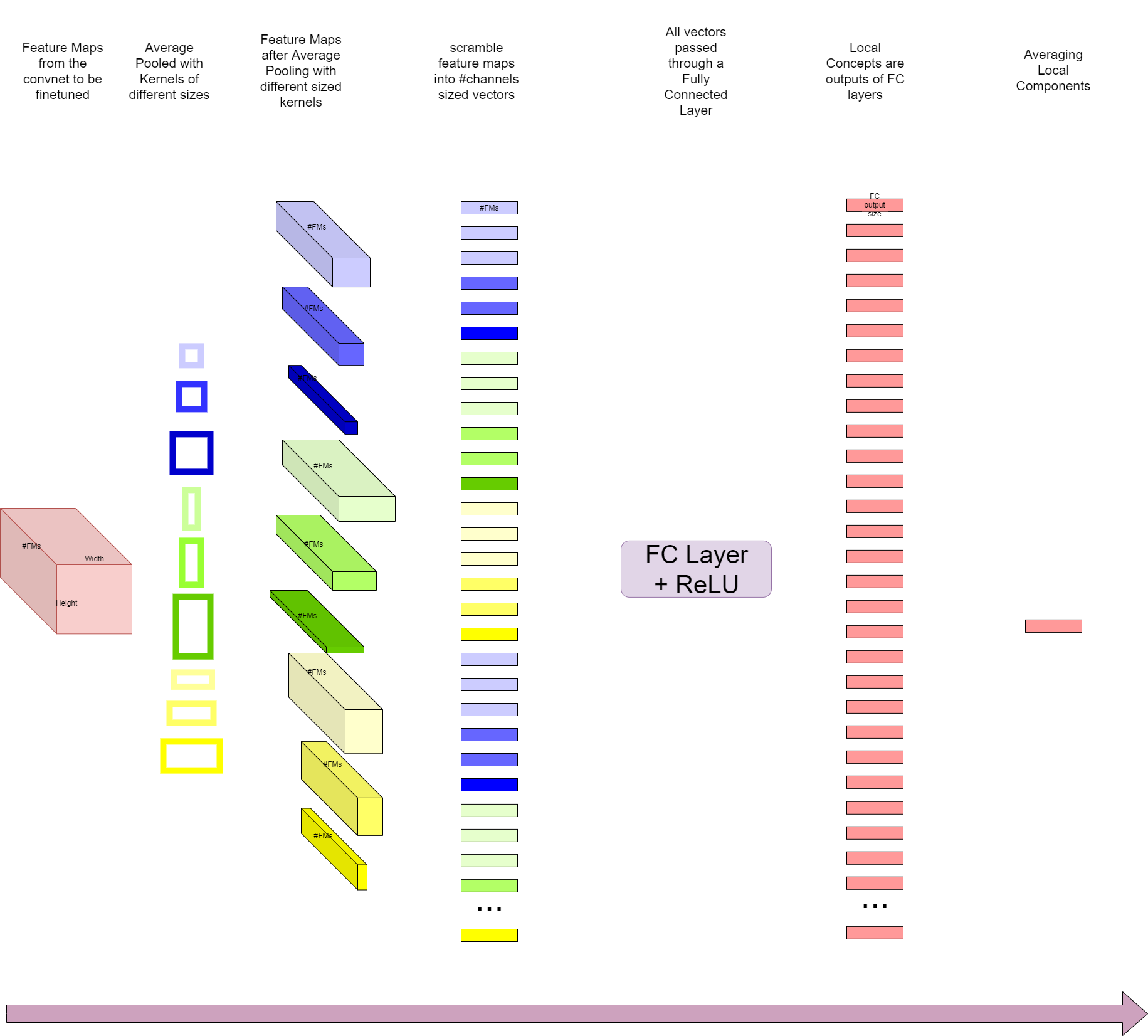}}
\par\end{centering}
\caption{This figure explains the implementation of Local Context Aggregation
Layer (LCA Layer). The feature maps of pretrained convnet (size \#FMs
X Height X Width) are averaged pooled by kernels of different sizes
and the pooled feature maps are then scrambled into vectors of \#FMs
size. These vectors are then passed through a Fully Connected layer
to give rise to different \textquotedblleft Local Concepts\textquotedblright{}
vectors. The Local Concepts Vectors are then aggregated by averaging
into the final vector for the image.}
\end{figure}

Implementation-wise, in a LCA layer, we build features for all possible
local concepts in an image and aggregate them by averaging. These
aggregated features are then fed to the classifier layer. This LCAlayer
can be used during finetuning any pretrained convnet when placed between
pretrained layers and classification layer. The implementation of
LCA layer can be visualized in Figure2.

\begin{figure}[htbp]
\begin{centering}
\includegraphics[width=1\linewidth]{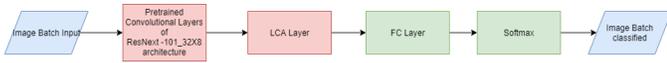}
\par\end{centering}
\caption{This figure shows the final Neural Network being finetuned for the
retail product image classification problem statement. The LCA layer
is placed between the final classification layer and pretrained convolutional
layers of ResNext.}
\end{figure}

For a resnext-101\_32X8 architecture, LCAlayer is placed between the
pretrained convolutional layers of the ResNext architecture and the
last FC classification layer. All possible rectangle and square kernels
larger than 1X1 are used to average pool the feature map from pretrained
network and get corresponding pooled feature maps. The number of feature
maps remains the same before and after each individual average pool
operation. Now all the pooled feature maps after average pool operations
are scrambled into vectors, each of dimension of number of feature
maps in ResNext pretrained output. Each of these vectors is passed
through a fully connected layer followed by Relu nonlinearity. Figure
3 shows arrangement of layers when finetuning the ResNext architecture
along with LSA layer. The Neural Network is trained with Stochastic
Gradient Descent with Momentum.

\subsection{Maximum Entropy loss as an auxiliary loss}

Maximum Entropy loss has been previously used for fine grained visual
classification\cite{6}. We show that using Maximum Entropy loss as
an auxiliary loss in retail product image classification loss betters
the accuracy of the convnet. This might be due to the fact that the
diversity of features in retail dataset is not as high as real world
objects and diversity of features within classes is not that high
too. Authors of \cite{6} propose that Maximum Entropy loss can be
useful in such circumstances. The structure of the loss function can
be found in Figure 4. 

\begin{figure}[htbp]
\begin{centering}
\textsf{\includegraphics[width=1\linewidth]{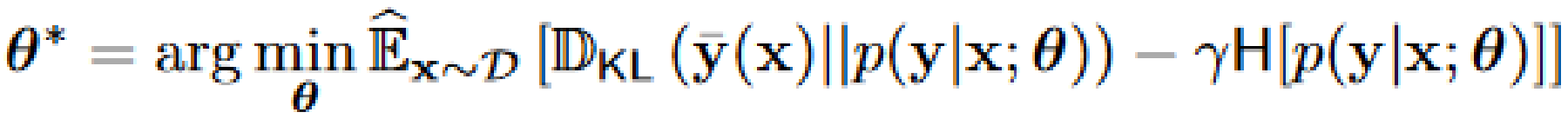}}
\par\end{centering}
\caption{The Maximum Entropy Loss Function}
\end{figure}

The description of Entropy (quantity H in equation of Maximum Entropy
Loss) over a conditional distribution can be found in Figure 5.

\begin{figure}[htbp]
\begin{centering}
\includegraphics[width=1\linewidth]{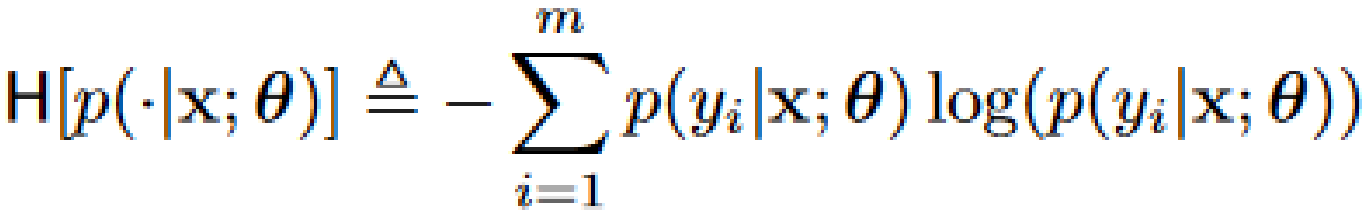}
\par\end{centering}
\caption{Entropy over a conditional probability distribution}
\end{figure}

A weighted average between Negative Likelihood loss and Maximum Entropy
loss is used as the final loss term for finetuning the neural networks.

\section{Datasets}

In this section, we describe the various datasets used for experiments.
We also explain how these datasets are analogues to real-world problem
statements.

\subsection{Baselines}

We choose two baselines to show the effectiveness of our method. The
first baseline is a simple finetuning\cite{5} of a ResNext\cite{2} model pretrained
on imagenet only (referred to as ResNext-INet). These baselines give
an idea about how much the tricks we implement one on the top of other,
aid classification accuracy. The other baseline is keypoint detection
and matching using convnet based Superpoint algorithm\cite{4}. This
is because keypoint matching based identification is common in many
retail product image classification systems. Also, Superpoint is one
of the leading methods for keypoint detection and feature matching.
This gives us a good upper bound on what retail product identification
systems using image matching could achieve.

\subsection{Grozi-120 dataset}

Grozi-120 dataset\cite{13} (available at \href{http://grozi.calit2.net/grozi.html}{link})
is a dataset having images of 120 retail products. Some products in
the dataset are for example : Cheerios, Cheez-it, Snickers etc. We
take in-vitro images of retail products as training data and in-situ
images were taken as testing data. Typically, the number of in-vitro
images is 4-6 per class. These in-vitro training images are packshots\cite{7}
taken from the internet and thus many augmentation techniques were
applied on the images before finetuning the convnet for classification.
Figure 6 shows a few pairs of in-vitro / in-situ images. In real world
use cases, such type of classification problem often comes up where
one gets only pack shots for training and the classifier trained has
to work on images from shops/retail outlets. 

\begin{figure}[htbp]
\begin{centering}
\textsf{\includegraphics[width=1\linewidth]{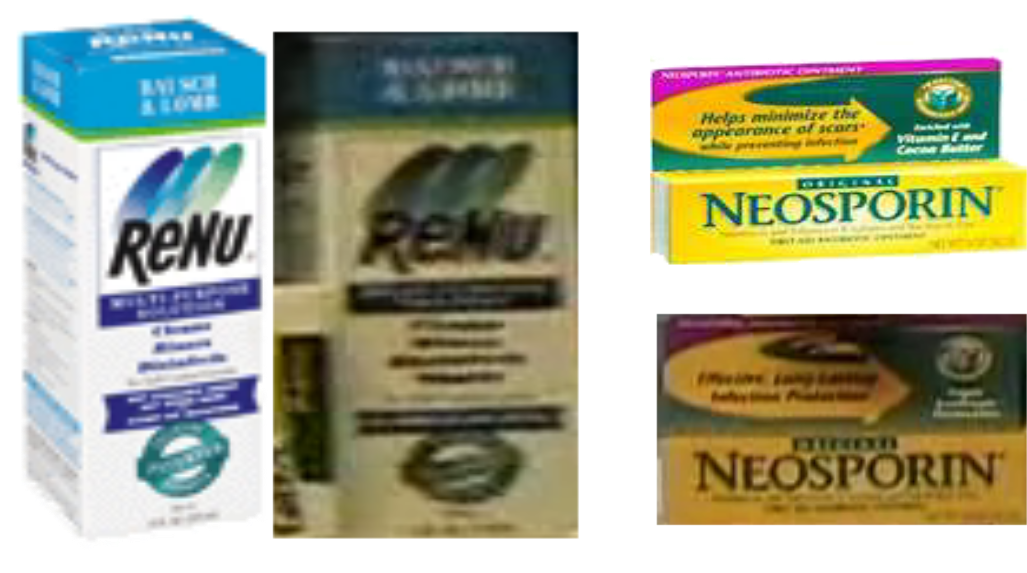}}
\par\end{centering}
\caption{Train and test pairs of images from Grozi-120 dataset.}
\end{figure}

\subsection{CAPG-GP dataset}

CAPG-GP\cite{14} (available at \href{\%20http://zju-capg.org/capg-gp.html}{link})
dataset has 102 retail products for fine grained one-shot classification.
All products have just one training image. However, the training images
are not pack shots but a small number of good quality images of actual
products. In real world, an analogous classification problem often
comes up where a few high quality images are available to train the
classifier and the classifier trained is supposed to work on product
images from shops/retail outlets. Image augmentations to incorporate different types of distortions into train set are introduced while the convnet is finetuned. Figure 7 shows a few pairs of train
and test set images.

\begin{figure}[htbp]
\begin{centering}
\includegraphics[width=1\linewidth]{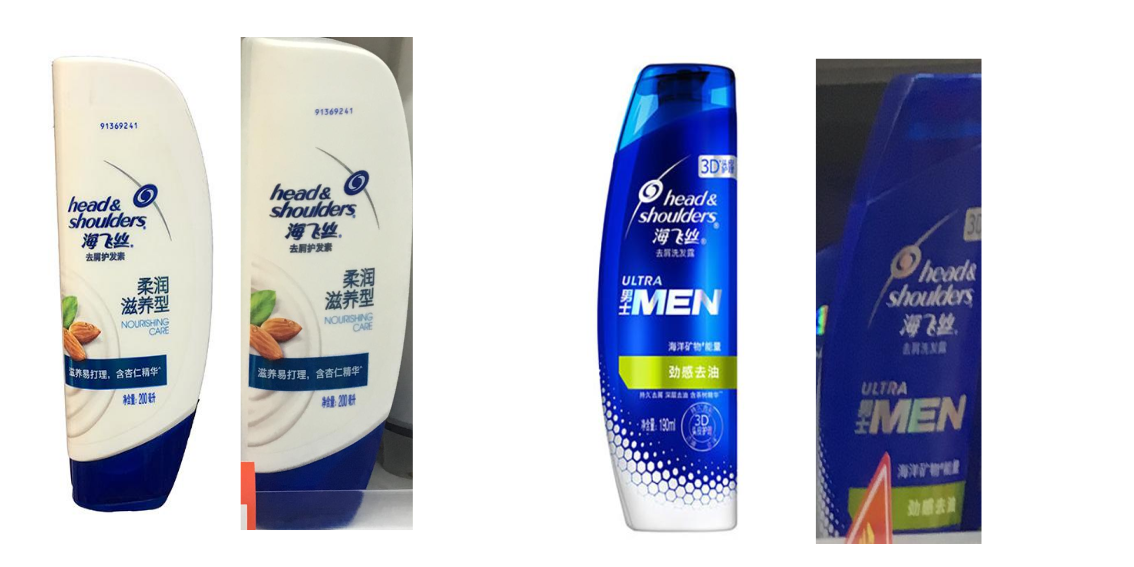}
\par\end{centering}
\caption{Train and Test pairs of images from CAPG-GP dataset.}
\end{figure}

\subsection{DM4VM dataset}

DM4VM dataset (available at \href{https://www.dropbox.com/s/n8zh37udjmrydad/20150320_DM4VM_dataset.zip?dl=0}{link})\cite{15}
is a dataset of 10 retail products with 60-70 images in training set
per product and approximately 30 images per product in the test set.
Both the training and test images appear to be real world images taken
from shelves. Figure 8 shows pairs of train and test set images. A
real world usecase analogous to this dataset is when data is collected
from real world shelves and is annotated in a considerable number
to train a classifier which again has to work in similar domain as
training data.

\begin{figure}[htbp]
\begin{centering}
\includegraphics[width=1\linewidth]{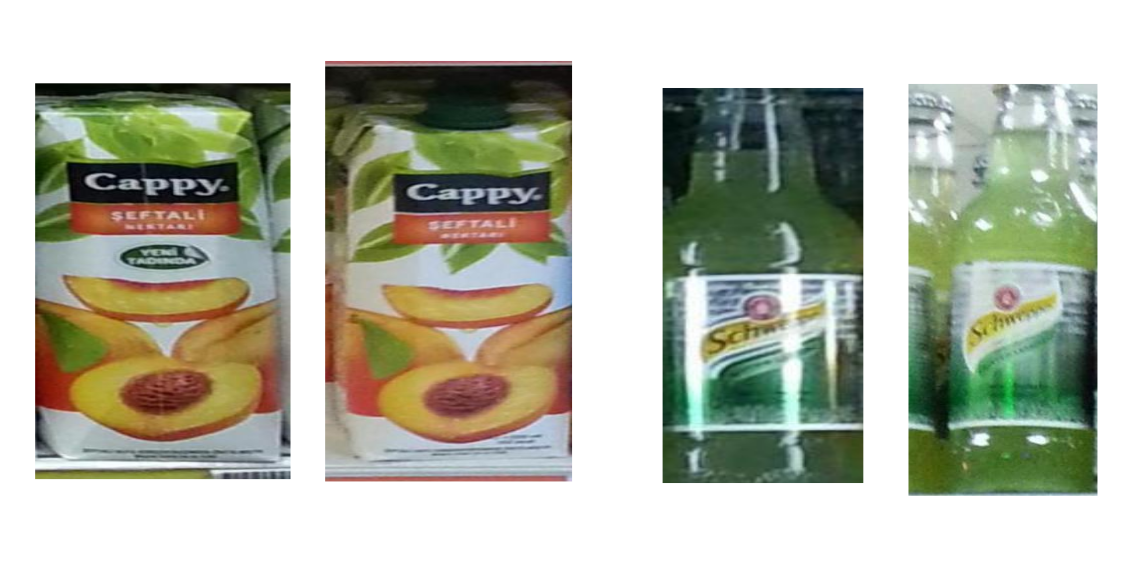}
\par\end{centering}
\caption{Train and Test pairs from DM4VM dataset.}
\end{figure}

\section{Results}

We now present the results of our experiments on various datasets
to show the accuracy gains the proposed tricks achieve. As mentioned
earlier, our baselines are classification by image matching (keypoint
detection + keypoint matching) using Superpoint\cite{4} and finetuning
Resnext\cite{2} pretrained on Imagenet (ResNext-INet). We then show
accuracy of our tricks added one on the top of the other. We show
results when finetuning ResNext-WSL\cite{8} on the retail product
classification datasets. When ResNext-WSL is trained along with addition
of the LCA layer, the model thus trained is referred to as ResNext-WSL-LCAlayer.
When Resnext-WSL with LCA layer is finetuned with a mutlitask learning
loss combining Negative Likelihood loss and Maximum-Entropy loss,
the model is called ResNext-WSL-LCAlayer-ME. The performance numbers
are test set classification accuracy in \%age.

\begin{table}[htbp]
\caption{Accuracy of various baselines and proposed methods}

\centering{}%
\begin{tabular}{|c|c|c|c|}
\hline 
Method / Dataset & Grozi-120 & CAPG-GP & DM4VM\tabularnewline
\hline 
Image Matching with Superpoint & 44.8\% & 84.7\% & 96.16\%\tabularnewline
\hline 
ResNext-INet & 58.66\%  & 83.9\% & 99.3\%\tabularnewline
\hline 
ResNext-WSL & 60.4\% & 84.1\% & 100\%\tabularnewline
\hline 
ResNext-WSL-LCAlayer & 70.8\% & 90.4\% & 100\%\tabularnewline
\hline 
ResNext-WSL-LCAlayer-ME & 72.3\% & 92.2\% & 100\%\tabularnewline
\hline 
\end{tabular}
\end{table}

When we analyze the results, we can come to a set of conclusions.
ResNext-WSL gets better accuracy than ResNext-INet for all datasets,
but its gets a relatively higher accuracy boost in Grozi-120, where
the test images have a lot of distortions and noise. We can attribute
this to the robustness pretraining it on instagram gives to the model.
Adding an LCAlayer gives a sizable accuracy boost in all the datasets
showing it is a good methodology for any type of retail product image
classification problem. Max-Entropy loss gives an accuracy boost across
all datasets too, reinforcing that the hypothesis of the inventors
that, it is a good add-on loss wherever low-diversity of features
is seen in the training data.

\section{Conclusions}

We propose multiple tricks that better the accuracy of retail product
image classification in multiple datasets. The technique of using
a instagram and later imagenet pretrained convnet instead of imagenet
pretrained convnet only is very simple to apply and gives performance boost
without adding any parameters. A new layer for neural networks is
proposed called LCA layer which, when added during finetuning, gives
consistent accuracy gain across all datasets. We also show that using
maximum-entropy loss as an auxiliary loss makes the classifier work better.

\appendices{}

\section*{Acknowlegment}

I thank my colleagues Srikrishna, Sonaal and Harshita for their help
in calculating baselines using keypoint matching as well as in the
implementation of maximum entropy loss.

\bibliographystyle{IEEEtran}
\bibliography{db}

\begin{thebibliography}{10}
\providecommand{\url}[1]{#1}
\csname url@samestyle\endcsname
\providecommand{\newblock}{\relax}
\providecommand{\bibinfo}[2]{#2}
\providecommand{\BIBentrySTDinterwordspacing}{\spaceskip=0pt\relax}
\providecommand{\BIBentryALTinterwordstretchfactor}{4}
\providecommand{\BIBentryALTinterwordspacing}{\spaceskip=\fontdimen2\font plus
\BIBentryALTinterwordstretchfactor\fontdimen3\font minus
  \fontdimen4\font\relax}
\providecommand{\BIBforeignlanguage}[2]{{%
\expandafter\ifx\csname l@#1\endcsname\relax
\typeout{** WARNING: IEEEtran.bst: No hyphenation pattern has been}%
\typeout{** loaded for the language `#1'. Using the pattern for}%
\typeout{** the default language instead.}%
\else
\language=\csname l@#1\endcsname
\fi
#2}}
\providecommand{\BIBdecl}{\relax}
\BIBdecl

\bibitem{1}
Y.~LeCun, Y.~Bengio, and G.~Hinton, ``Deep learning,'' \emph{nature}, vol. 521,
  no. 7553, p. 436, 2015.

\bibitem{2}
S.~Xie, R.~Girshick, P.~Doll{\'a}r, Z.~Tu, and K.~He, ``Aggregated residual
  transformations for deep neural networks,'' in \emph{Proceedings of the IEEE
  conference on computer vision and pattern recognition}, 2017, pp. 1492--1500.

\bibitem{3}
E.~Goldman, R.~Herzig, A.~Eisenschtat, J.~Goldberger, and T.~Hassner, ``Precise
  detection in densely packed scenes,'' in \emph{Proceedings of the IEEE
  Conference on Computer Vision and Pattern Recognition}, 2019, pp. 5227--5236.

\bibitem{4}
D.~DeTone, T.~Malisiewicz, and A.~Rabinovich, ``Superpoint: Self-supervised
  interest point detection and description,'' in \emph{Proceedings of the IEEE
  Conference on Computer Vision and Pattern Recognition Workshops}, 2018, pp.
  224--236.

\bibitem{7}
\BIBentryALTinterwordspacing
Packshot wikipedia entry. [Online]. Available:
  \url{https://en.wikipedia.org/wiki/Packshot}
\BIBentrySTDinterwordspacing

\bibitem{5}
J.~Yosinski, J.~Clune, Y.~Bengio, and H.~Lipson, ``How transferable are
  features in deep neural networks?'' in \emph{Advances in neural information
  processing systems}, 2014, pp. 3320--3328.

\bibitem{6}
A.~Dubey, O.~Gupta, R.~Raskar, and N.~Naik, ``Maximum-entropy fine grained
  classification,'' in \emph{Advances in Neural Information Processing
  Systems}, 2018, pp. 637--647.

\bibitem{8}
D.~Mahajan, R.~Girshick, V.~Ramanathan, K.~He, M.~Paluri, Y.~Li, A.~Bharambe,
  and L.~van~der Maaten, ``Exploring the limits of weakly supervised
  pretraining,'' in \emph{Proceedings of the European Conference on Computer
  Vision (ECCV)}, 2018, pp. 181--196.

\bibitem{17}
D.~G. Lowe, ``Distinctive image features from scale-invariant keypoints,''
  \emph{International journal of computer vision}, vol.~60, no.~2, pp. 91--110,
  2004.

\bibitem{18}
C.~G. Harris, M.~Stephens \emph{et~al.}, ``A combined corner and edge
  detector.'' in \emph{Alvey vision conference}, vol.~15, no.~50.\hskip 1em
  plus 0.5em minus 0.4em\relax Citeseer, 1988, pp. 10--5244.

\bibitem{16}
I.~Baz, E.~Yoruk, and M.~Cetin, ``Context-aware hybrid classification system
  for fine-grained retail product recognition,'' in \emph{2016 IEEE 12th Image,
  Video, and Multidimensional Signal Processing Workshop (IVMSP)}.\hskip 1em
  plus 0.5em minus 0.4em\relax IEEE, 2016, pp. 1--5.

\bibitem{13}
M.~Merler, C.~Galleguillos, and S.~Belongie, ``Recognizing groceries in situ
  using in vitro training data,'' in \emph{2007 IEEE Conference on Computer
  Vision and Pattern Recognition}.\hskip 1em plus 0.5em minus 0.4em\relax IEEE,
  2007, pp. 1--8.

\bibitem{14}
W.~Geng, F.~Han, J.~Lin, L.~Zhu, J.~Bai, S.~Wang, L.~He, Q.~Xiao, and Z.~Lai,
  ``Fine-grained grocery product recognition by one-shot learning,'' in
  \emph{2018 ACM Multimedia Conference on Multimedia Conference}.\hskip 1em
  plus 0.5em minus 0.4em\relax ACM, 2018, pp. 1706--1714.

\bibitem{23}
S.~Leutenegger, M.~Chli, and R.~Siegwart, ``Brisk: Binary robust invariant
  scalable keypoints,'' in \emph{2011 IEEE international conference on computer
  vision (ICCV)}.\hskip 1em plus 0.5em minus 0.4em\relax Ieee, 2011, pp.
  2548--2555.

\bibitem{19}
A.~Franco, D.~Maltoni, and S.~Papi, ``Grocery product detection and
  recognition,'' \emph{Expert Syst. Appl.}, vol.~81, pp. 163--176, 2017.

\bibitem{20}
M.~George and C.~Floerkemeier, ``Recognizing products: A per-exemplar
  multi-label image classification approach,'' in \emph{ECCV}, 2014.

\bibitem{21}
L.~Karlinsky, J.~Shtok, Y.~Tzur, and A.~Tzadok, ``Fine-grained recognition of
  thousands of object categories with single-example training,'' in
  \emph{Proceedings of the IEEE Conference on Computer Vision and Pattern
  Recognition}, 2017, pp. 4113--4122.

\bibitem{22}
A.~Tonioni and L.~di~Stefano, ``Product recognition in store shelves as a
  sub-graph isomorphism problem,'' in \emph{ICIAP}, 2017.

\bibitem{24}
A.~Tonioni and L.~Di~Stefano, ``Domain invariant hierarchical embedding for
  grocery products recognition,'' \emph{Computer Vision and Image
  Understanding}, vol. 182, pp. 81--92, 2019.

\bibitem{25}
S.~Varadarajan, S.~Kant, and M.~M. Srivastava, ``Benchmark for generic product
  detection: A low data baseline for dense object detection,'' 2019.

\bibitem{10}
O.~Russakovsky, J.~Deng, H.~Su, J.~Krause, S.~Satheesh, S.~Ma, Z.~Huang,
  A.~Karpathy, A.~Khosla, M.~Bernstein \emph{et~al.}, ``Imagenet large scale
  visual recognition challenge,'' \emph{International journal of computer
  vision}, vol. 115, no.~3, pp. 211--252, 2015.

\bibitem{9}
\BIBentryALTinterwordspacing
Instagram wikipedia entry. [Online]. Available:
  \url{https://en.wikipedia.org/wiki/Instagram}
\BIBentrySTDinterwordspacing

\bibitem{11}
A.~E. Orhan, ``Robustness properties of facebook's resnext wsl models,''
  \emph{arXiv preprint arXiv:1907.07640}, 2019.

\bibitem{12}
A.~Paszke, S.~Gross, F.~Massa, A.~Lerer, J.~Bradbury, G.~Chanan, T.~Killeen,
  Z.~Lin, N.~Gimelshein, L.~Antiga \emph{et~al.}, ``Pytorch: An imperative
  style, high-performance deep learning library,'' in \emph{Advances in Neural
  Information Processing Systems}, 2019, pp. 8024--8035.

\bibitem{15}
\BIBentryALTinterwordspacing
C.~B. Akgul. Color histogram descriptors, data mining for visual media 2015,
  assignment 04. [Online]. Available:
  \url{http://www.cba-research.com/pdfs/DM4VM_A04_ColorDescriptors.pdf}
\BIBentrySTDinterwordspacing

\end{thebibliography}

\end{document}